\documentclass[10pt,twocolumn,letterpaper]{article}
\usepackage{amsmath}
\usepackage{cvpr}
\usepackage{times}
\usepackage{epsfig}
\usepackage{graphicx}
\usepackage{amsmath}
\usepackage{amssymb}
\usepackage{multirow}
\usepackage{array}
\usepackage{booktabs}       
\usepackage{diagbox}
\usepackage{array}
\usepackage{tabularx}
\usepackage{multirow}
\usepackage{makecell}
\usepackage{latexsym}
\usepackage{arydshln}
\usepackage{bm}
\usepackage{subfigure}
\usepackage{xcolor}
\usepackage{pmat}
\usepackage[misc]{ifsym}
\usepackage{bibspacing}
\newcolumntype{M}[1]{>{\centering\arraybackslash}m{#1}}
\usepackage[colorlinks=true,breaklinks=true,bookmarks=false,citecolor=blue]{hyperref}

\cvprfinalcopy 

\def\cvprPaperID{****} 

\begin{document}

\title{Short Range Correlation Transformer for Occluded Person Re-Identification}

\author{Yunbin Zhao, Songhao Zhu$^*$, Dongsheng Wang, Zhiwei Liang\\
College of Automation and Artificial Intelligence, \\Nanjing University of Posts and Telecommunications, Nanjing, China\\zhush@njupt.edu.cn
}

\maketitle

\begin{abstract}
    Occluded person re-identification is one of the challenging areas of computer vision, which faces problems such as inefficient feature representation and low recognition accuracy.
    Convolutional neural network pays more attention to the extraction of local features, therefore it is difficult to extract features of occluded pedestrians and the effect is not so satisfied.
    Recently, vision transformer is introduced into the field of re-identification and achieves the most advanced results by constructing the relationship of global features between patch sequences.
    However, the performance of vision transformer in extracting local features is inferior to that of convolutional neural network. 
    Therefore, we design a partial feature transformer-based person re-identification framework named PFT. The proposed PFT utilizes three modules to enhance the efficiency of vision transformer.
    (1) Patch full dimension enhancement module. We design a learnable tensor with the same size as patch sequences, which is full-dimensional and deeply embedded in patch sequences to enrich the diversity of training samples.
    (2) Fusion and reconstruction module. 
    We extract the less important part of obtained patch sequences, and fuse them with original patch sequence to reconstruct the original patch sequences.
    (3) Spatial Slicing Module. We slice and group patch sequences from spatial direction, which can effectively improve the short-range correlation of patch sequences. 
    Experimental results over occluded and holistic re-identification datasets demonstrate that the proposed PFT network achieves superior performance consistently and outperforms the state-of-the-art methods.
\end{abstract}

\section{Introduction}
Person re-identification\cite{1} aims at linking target persons in different cameras, and is widely used in security, surveillance and other fields.
In recent years, a large number of methods\cite{2,3,4,5,6} for solving re-identification have been proposed. Most of these methods are based on convolutional neural networks to extract human features, and have achieved satisfactory results on mainstream datasets. However, in real life, we often encounter situations such as incomplete or occluded person and cluttered backgrounds. In such cases, most person re-identification methods are difficult to achieve satisfied recognition accuracy.

The receptive field of convolutional neural network is limited to a small area by the Gaussian distribution\cite{7}. Existence of a large amount of occlusion information, background information or other noise\cite{17,8}, the small receptive field will easily ignore important characteristic information. Furthermore, the down-sampling operation in convolutional neural network will reduce the resolution of feature mAP, which leading to the decline of its identification performance. Therefore, even if the attention mechanism\cite{13,14,15,16} or feature alignment methods\cite{9} are introduced , it is difficult to solve the challenge of re-identification of occluded person.

Vision transformer\cite{10} has been proved to have good performance in image classification, and the effectiveness is no less than traditional convolutional neural network method. Vision transformer takes multi-head self-attention mechanism\cite{11} as the core, and discards convolution and down-sampling operations. Specifically, each original image is first cut into patch sequence and input into the network, then patch sequence is embedded with class coding and position coding, and finally self-attention is performed on patch sequence. Recently, vision transformer has been introduced into the field of person re-identification. With its ability of capturing global features and application of self-attention mechanism, vision transformer achieves the best effectiveness at the time, and the same is true for re-identification of occluded person.

With a good long-range correlation to long sequence, vision transformer has achieved good results in person re-identification. However, it is also prone to misjudgment when the person is occluded heavily or the background is similar to the person. It can be seen that vision transformer is not good at capturing local features\cite{12} in the case of occlusion, resulting in poor robustness.
\begin{figure}[t!]
    \centering
    \includegraphics[width=1.0\linewidth,height=90pt]{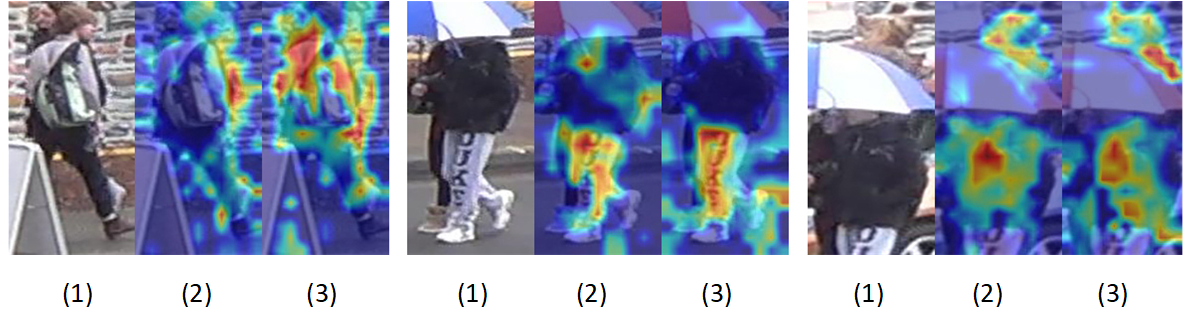}
    \caption{Grad-CAM\cite{18} attention maps of differnet transformer:(1)Original images,(2)Vision Transformer-based methods,(3)Ours PFT can pay more attention on local feature and have a bigger feeling field} 
    \label{fig:2.png}
\end{figure}
Therefore, we propose a PFT with three modules based on vision transformer to improve the short-range correlation of patch sequence and extract local features of non-occluded part. It can be seen from Figure \ref{fig:2.png} that PFT pays more attention to local features than original vision transformer. Next, we will detail the proposed three modules.

Firstly, we propose a patch sequence reconstruction module (FRM) to fuse the noise patch or background patch with object patch to reduce the influence of noise or occlusion on the overall patch sequence, and then reconstruct the whole original patch sequence. In this way, the receptive field of reconstructed patch sequence can obtain more local features since the fusion operation helps to enlarge the  proportion of object feature in global feature.

Secondly, to improve the generalization and robustness of vision transformer, we propose a patch sequence spatial slicing module (SSM) to slice the patch sequence in the spatial direction after the last layer of vision transformer. As a global branch with spatial correlation of patch sequences, the module can lengthen the distance between fine categorization and improve the short-range correlation of sequences, therefore the network pays more attention to local features and has stronger generalization ability.

Finally, many workers pay attention to the circulation of patch sequence and ignore the enhanceability of patch sequence itself. Therefore, we introduce the patch full dimension sequence enhancement (PFDE) coding into the patch cutting module. The coding is a learnable tensor, which can reduce the noise during the image input stage and extract more discriminative features of occluded person. Therefore, PFDE helps to reduce the large noise of input images and the difficulty of extracting discriminative features.

The main contributions of our works are described as follows:
\begin{itemize}
    \item A patch full-dimension enhancement module of patch sequence is designed to increase the distance between fine categorizations within a reasonable range, enrich the diversity of training samples, weaken the noise and highlight the discriminative features.
    \item A fusion and reconstruction module of patch sequence is proposed to widen the differentiation of feature representations between fine categorizations, improve the proportion of person recognizable features in patch sequence, and enhance the generalization and robustness of the reconstructed patch sequence.
    \item A spatial slicing module is designed to extract distinguishable features of patch sequence from the spatial direction. Furthermore, this module integrates the spatial correlation of input images into patch sequence to improve the short-range correlation of patch sequence, which helps to extract comprehensive local features of occluded person to improve the generalization performance under different occlusion conditions.
\end{itemize}

\section{Related Work}
\textbf Most of the researches on person re-identification focus on complete person image, and rarely consider occluded person image. However, in real life, re-identification of occluded person cannot be ignored. Especially in crowded scenes, it is difficult to obtain a complete person image most of the time. Therefore, let’s review a few items about occluded re-identification.
\begin{figure*}[t]
    \centering
    \includegraphics[width=1.0\linewidth]{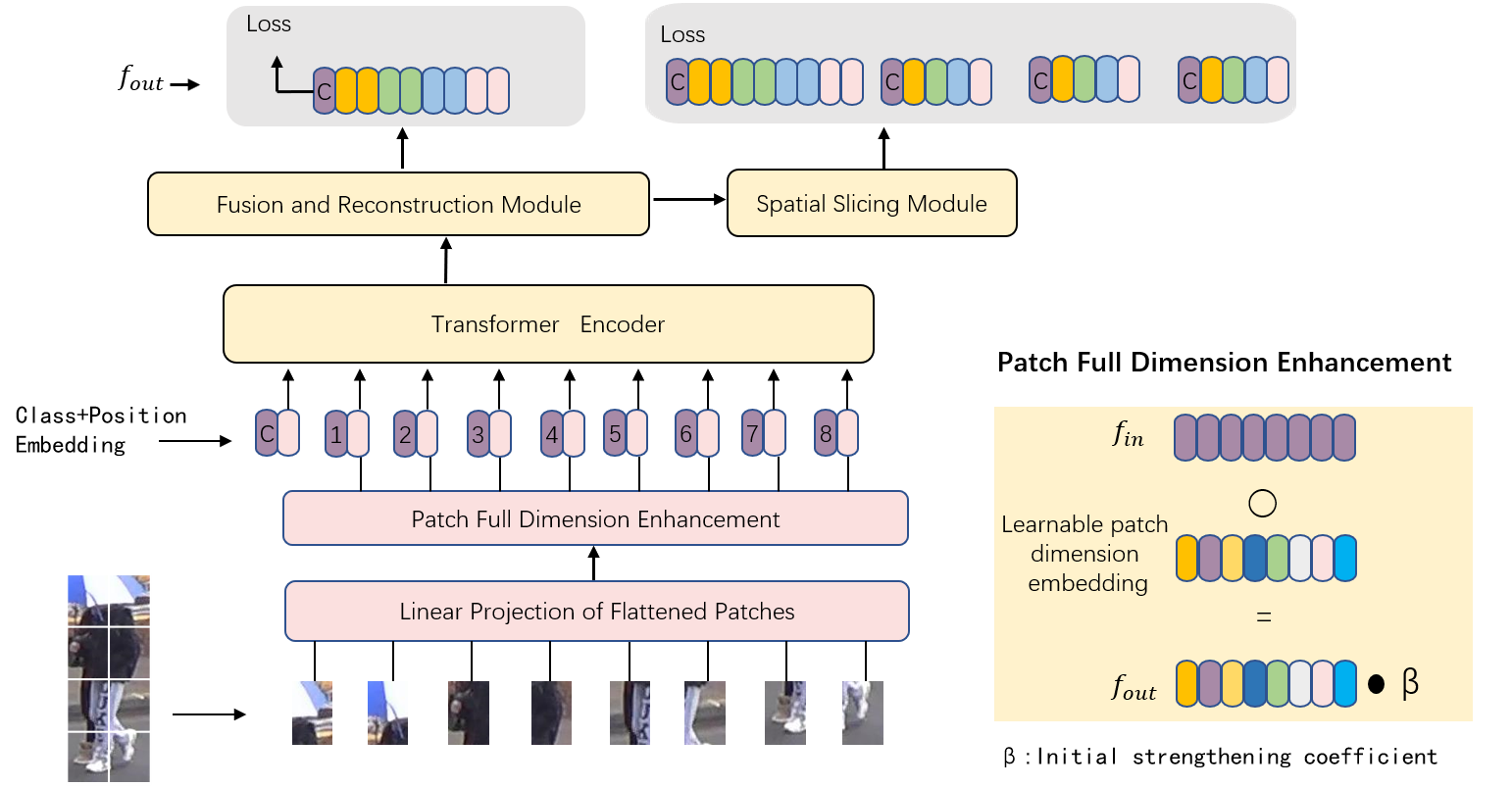}
    \caption{Framework of proposed PFT, which contains three new modules, PFDE, FRM and SSM according to the structural order of the network. The structure of PFDE is illustrated in detail, and the detailed structure of other modules is shown in Fig. 3 and Fig. 4} 
    \label{fig:Overview}
\end{figure*}

\textbf{Occluded Person ReID.}
Existing deep learning methods for re-identification of occluded persons are mainly based on convolutional neural network. The main design ideas are feature alignment or the introduction of high-order semantic information, where high order semantic information refers to attitude guidance information\cite{19}. Key points of human body are first estimated, and then the pose is used to identify the occluded person. CGEA layer is proposed to jointly learn and embed the attitude guidance information of local features, and directly predict the similarity score. It not only makes full use of the alignment learned by graph matching, but also uses robust soft matching instead of sensitive one-to-one matching. Gao\cite{22} et al. propose a posture-guided visual part matching method, which uses posture-guided attention to jointly learn distinguishing features and self-digs the visibility of parts in an end-to-end framework. Although the introduction of attitude guidance information makes the model have higher recognition rate than before, the additional key point estimation model makes the overall model more bloated and reduces the running speed of the network.

\textbf{Vision Transformer}
Transformer is a common model in NLP\cite{23,24,25,26} field. Ashish Vaswani\cite{10} et al. propose multi-head self-attention mechanism, completely abandon network structures such as RNN and CNN, and only use self-attention for machine translation tasks, and achieve good results. Google and other authors introduce transformer into the field of image classification, propose vision transformer, segment the image into sequences and input it into transformer, preserve the original structure of transformer to the greatest extent, and achieve very good results. However, vision transformer needs a large number of datasets for pre-training to obtain training results similar to CNN. Therefore, touvron et al.\cite{20} propose the Deit framework and optimize the problem by using teacher student strategy. Recently, Shuting he et al.\cite{21} propose Transreid and apply the vision transformer to the field of object re-identification. They also propose the JPM module to shuffle and classify the features of the last layer of the network and then calculate their losses respectively, which further enhance the learning of robust features by Transreid. However, Transreid still focuses more on learning global features, and the impact of local occlusion features and short-range dependence have not been well solved.

\section{Methodology}
Our PFT network for occluded person re-identification is based on original vision transformer network. However, to improve the ability of self-learning local features and enhance short-range correlation, we design a patch full dimension embedding module in the patch embedding stage to optimize the patch embedding operation. At the same time, we also propose a fusion and reconstruction module and a spatial slicing module to fuse local features and extract local features of feature mAP in the spatial direction, and enhance the robustness of feature extraction.

\begin{figure*}[t]
    \centering
    \includegraphics[width=1.0\linewidth]{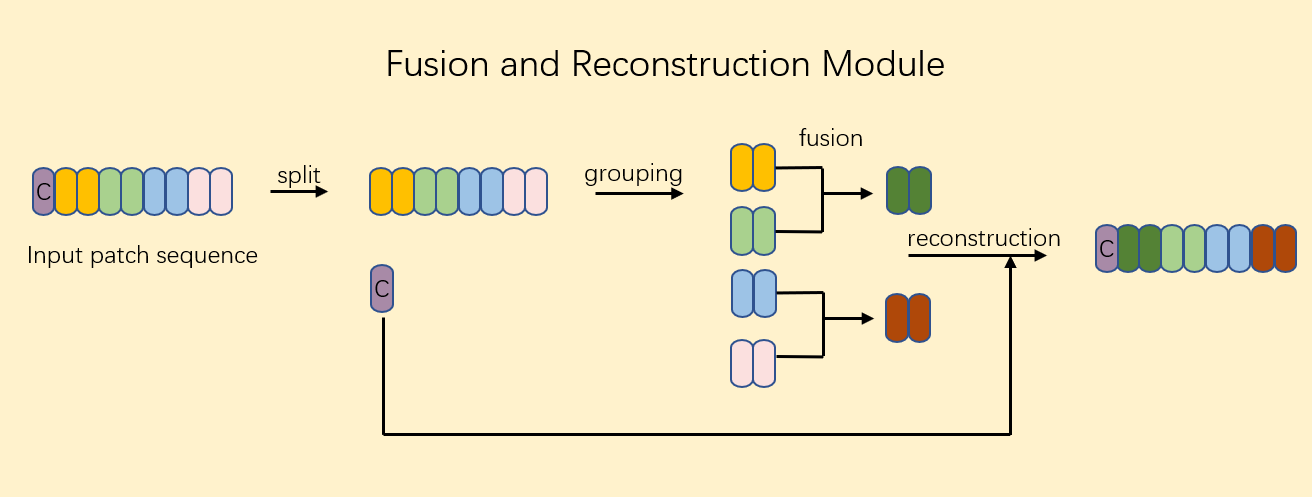}
    \caption{Illustration of the proposed FRM. Split is to embed the classification into the segmentation operation, grouping is to group the original patch sequence, fusion is to add and fuse the selected patches, and reconstruction is to reconstruct the new patch sequence.} 
    \label{fig:FRM2}
\end{figure*}
\subsection{Patch Full Dimension Enhancement}
 
We propose a learnable patch full dimension tensor to enhance the patch embedding operation. PFDE is depicted in Figure \ref{fig:Overview}, the given input image $x\in R^{H\times W\times C}$, where H, W and C represent the height, width and channel dimensions of the input image respectively. The size of input image is 256${\times}$128. After the patch embedding operation, the input image is divided into N patch blocks of the same size.

At this time, the input image size is changed from ${batchsize\times H\times W\times C}$ to ${batchsize \times N\times D}$, that is, the RGB image is converted into a two-dimensional patch sequence. Due to the convolution operation in patch embedding, the resolution of occluded person image is not high. Under the influence of translation invariance of convolution operation, the original semantic information will be offset, which will deeply affect the information in original input image from the channel direction. Therefore, we construct a learnable tensor with the same size as the patch sequence after convolution operation, $LPDE=N\times D$, where N is the number of patches and D is the dimension of each patch. The input patch sequence is $f_{in}$, where H and W are the height and width of the input image respectively, P is the side length of each patch, and S is the stride size.
\begin{small}\begin{equation}
\begin{aligned}
    &f_{in} = {[f_1;f_2;f_3;f_4;f_5...f_N]},\\
    &N = N_H\times N_W = [\frac{H+S-P}{S}]\times [\frac{W+S-P}{S}]
\end{aligned}
\end{equation}\end{small}

After constructing a learnable tensor $LPDE$ with the same size as the input patch sequence, and making Hadamard product between $LPDE$ and input patch sequence, we obtain the output patch sequence $f_{out}$.
\begin{small}\begin{equation}
\begin{aligned}
    &LPDE = {[p_1;p_2;p_3;p_4;p_5...p_N]},\\
    &f_{out} = [f_1p_1;f_2p_2;f_3p_3;f_4p_4;...f_NP_N]
\end{aligned}
\end{equation}\end{small}
As a learnable tensor, $LPDE$ is initialized with all ones and embedded into the input patch sequence through Hadamard product operation to enhance the feature expression ability of the input patch sequence in the training process. 
Moreover, after pre-training, vision transformer can accelerate the convergence speed of the PFT network, and the optimization speed of self-learning is far lower than the convergence speed of the network itself. 
Therefore, the patch full-dimensional enhancement module will not have a negative impact on the network, and improve network optimization capabilities within a certain range. With low resolution of input images, the module can introduce additional feature information into the training process to enrich the diversity of training samples. Its self-learning ability can improve the identification accuracy of the network, optimize the feature representation of input image, strengthen the contextual cues of occluded person and weaken the noise information.


\subsection{Fusion and Reconstruction Module}

\begin{figure*}[t]
    \centering
    \includegraphics[width=1.0\linewidth]{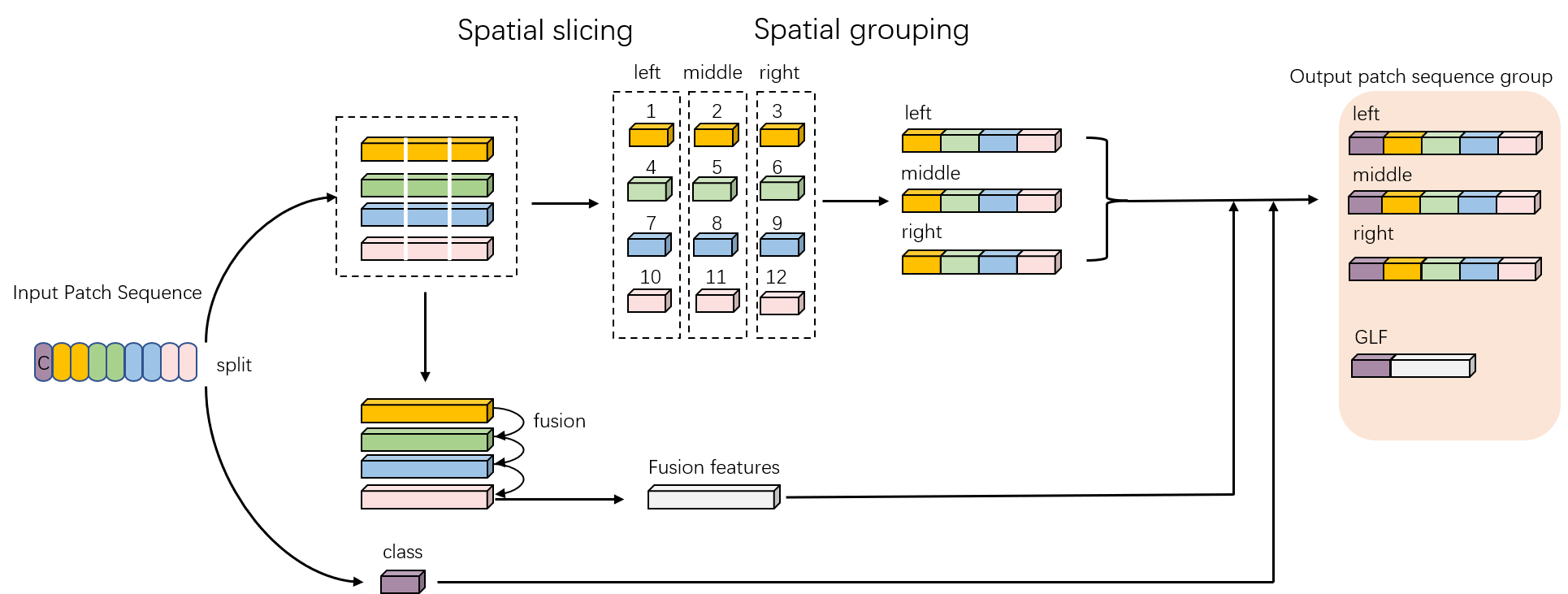}
    \caption{Overview of SSM} 
    \label{fig:SCM}
\end{figure*}

Although vision transformer can make good use of global features to achieve effective identification performance, the key information in the re-identification of occluded person usually depends more on local features. 

Since there are a lot of background, occlusion and other noise information in the task of occluded person re-identification, therefore patches with more occluded feature information are considered as more important ones in the patch sequence. The cosine similarity between different patches is calculated, and it is found that the information related to occluded person is mainly concentrated in the middle of the sequence. 
Because the similarity between the first or last patch and the global patch is very low, the probability of occasional high similarity is also very low, which indicates that the first and last patch are not important patches.
To this end, we will access the FRM module at the last layer, as shown in Figure \ref{fig:FRM2}. The input patch sequence of module FRM is  ${Z_{in}=[Class;z_1,z_2,z_3...z_N]}$.
\begin{itemize}
    \item Split: First, The $Class$ token is separated from the patch sequence $Z_{in}$. Then, new patch sequence ${F=[z_1,z_2,z_3...z_N]}$ is obtained.
    \item Grouping: Patch sequence $F$ is divided into four groups according to the same length, so four new partial patch sequences are achieved.
        \begin{equation}\label{split}
        \begin{aligned}
            &F_1=[z_1,z_2,z_3...z_{\frac{N}{4}}], \\       &F_2=[z_{\frac{N+1}{4}},z_{\frac{N+2}{4}},z_{\frac{N+3}{4}}...z_{\frac{2N}{4}}]\\
            &F_3=[z_{\frac{2N+1}{4}},z_{\frac{2N+2}{4}},z_{\frac{2N+3}{4}}...z_{\frac{3N}{4}}]\\
            &F_4=[z_{\frac{3N+1}{4}},z_{\frac{3N+2}{4}},z_{\frac{3N+3}{4}}...z_N]
        \end{aligned}
        \end{equation}
    \item Fusion: It is found that the correlation and dependence of head and tail parts in the patch sequence are low, but we don't want to completely discard these patches. 
    Although most of the time, the head and tail patches do not contain some important distinguishing features, sometimes they contain auxiliary distinguishing features such as head, hat, shoes and umbrella.
    So we want to fuse the features of the head and tail patches and replace them as $NewF_1$ and $NewF_4$.
        \begin{equation}
        \begin{aligned}
        \label{split3}
            &NewF_1=F_1+F_2 \\ 
            &=[z_1+z_{\frac{N+1}{4}},z_2+z_{\frac{N+2}{4}}...,z_{\frac{N}{4}}+z_{\frac{2N}{4}}] \\
            &=[m_1,m_2,...m_{\frac{N}{4}}] \\
            &NewF_4=F_3+F_4 \\
            &=[z_{\frac{2N+1}{4}}+z_{\frac{3N+1}{4}},z_{\frac{2N+2}{4}}+z_{\frac{3N+2}{4}}...z_{\frac{3N}{4}}+z_N] \\
            &=[L_1,L_2,..L_N] 
        \end{aligned}
        \end{equation}
    \item Reconstruction: After the new ${F_1}$ and ${F_4}$ are obtained, the four patch sequences of new ${F_1}$, ${F_2}$, ${F_3}$ and new ${F_4}$ are spliced into the patch sequence of the original size according to the original order. Therefore, the output of FRM is shown in the following formula \ref{split2}.
        \begin{equation}\label{split2}
        \begin{aligned}
            Z_{out} = [Class,NewF_1,F_2,F_3,NewF_4]
        \end{aligned}
        \end{equation}
\end{itemize}

\subsection{Spatial Slicing Module}

Vision transformer mainly focuses on global features from a linear perspective, so as to obtain the global correlation between patches. 
From the perspective of image processing, an image can be considered as a combination of two-dimensional patch sequences. Therefore, patch sequence still has spatial correlation.
In other words, not only successive patches have strong spatial correlation, but patches far apart also have certain spatial correlation. Therefore, we try to explore the spatial correlation of patch sequence to make up for the short correlation of patch sequence and ensure receptive field pays more attention to local features.

Inspired by jigsaw patch module\cite{21}, we follow its structure of deriving branches from the last layer of vision transformer, and introduce our designed SSM module to extract spatial correlation features of patch sequences. As shown in Figure \ref{fig:SCM}.

Firstly, patch sequence is separated from the class token by split operation and is divided into four groups, as shown in formula \ref{split}.
\begin{itemize}
    \item \setlength{\itemsep}{0pt}Spatial slicing: The obtained four groups patch sequences are sliced twice to the same length from the spatial direction. In this way, a total of 12 groups of different local patch sequences will be generated, and the corresponding numbers are assigned to the 12 groups of patch sequences, as shown in formula
    \ref{split4}.
        \begin{small}
        \begin{equation}\label{split4}
        \begin{aligned}
            &F_g = [z_m,...z_{m+\frac{N}{12}}] \quad g\epsilon [1,12]
            \\ &m\epsilon [1,\frac{N+1}{12},\frac{2N+1}{12},...\frac{11N+1}{12}]
        \end{aligned}
        \end{equation}
        \end{small}
    \item Spatial grouping: For obtained 12 sets of patch sequences, patches numbered 1, 4, 7 and 10 are spliced into left patch sequences, patches numbered 2, 5, 8 and 11 are spliced into middle patch sequences, and patches numbered 3, 6, 9 and 12 are spliced into right patch sequences.
    \item Fusion: The features of original four groups of patch sequences are fused to obtain fused patch sequence: Fusion features. The fused features have global feature information, and enlarge the distance between different targets.
    \item Splicing: The initial class token is spliced with left, middle, right and fusion features respectively. After fusion feature is spliced with the class token, it has local and global features, which is called Global and Local Features(GLF) as shown in formula \ref{split5}.
        \begin{small}
        \begin{equation}\label{split5}
        \begin{aligned}
            &GLF = [class,F_{1}+F_{2}+F_{3}+F_{4}] 
        \end{aligned}
        \end{equation}
        \end{small}
\end{itemize}
    
    Then, four new patch sequences namely left, middle, right and  GLF are obtained, as shown in the formula \ref{split6}.
        \begin{small}
        \begin{equation}\label{split6}
        \begin{aligned}
            &Out = [left,middle,right,GLF] 
        \end{aligned}
        \end{equation}
        \end{small}
\section{Experiments}
In this part, we conduct a comprehensive experiment on the proposed PFT based on vision transformer to test its effectiveness in enhancing the short-range correlation and long-range correlation of patch sequence in the field of re-identification of occluded person.

\subsection{Experimental Datasets}
We evaluate our proposed method on four person
ReID datasets, Occluded-ReID\cite{8}, Occluded-Duke\cite{9}, Market-1501\cite{27}, DukeMTMC-reID\cite{28}, Partial-REID\cite{29}, Partial-iLIDS\cite{47}.

\textbf{Occluded-Duke.} It contains 15,618 training images, 17,661 gallery images, and 2,210 occluded query images, which is by far the largest occluded re-ID datasets.

\textbf{Occluded-ReID.} Images are captured by mobile
camera equipments in campus, including 2000 annotated
images belonging to 200 identities. Among the dataset,
each person consists of 5 full-body person images and 5
occluded person images with various occlusions.

\textbf{Market-1501.} It consists of 32,668 images
of 1,501 identities captured by 6 camera views. Following
the standard setting, the whole dataset is divided into
a training set containing 12,936 images of 751 identities
and a testing set containing 19,732 images of 750 identities.

\textbf{DukeMTMC-reID.} It contains of 36,411 images of 1,812
persons from 8 cameras. 16,522 images of 702 persons are
randomly selected from the dataset as the training set, and the
remaining images are divided into the testing set containing
2,228 query images and 17,661 gallery images

\textbf{Partial-REID.} It is the first dataset for partial person re-identification, which includes 900 images of 60 persons, with 5 full-body person images, 5 partial person images and 5 occluded person images each identity. The images are collected at a university campus with various viewpoints and occlusions.

\textbf{Partial-iLIDS.} It is a simulated partial person Re-ID
dataset based on the iLIDS dataset, which has a total of 476 images
of 119 people.

\subsection{Implementation Details}
\textbf{Backbone.} We utilize vision transformer as the basic backbone network to cut input image into patch sequences, and embed class token and position coding for person re-identification. Then, we add the proposed PFDE, FRM and SSM module to construct a new framework named PFT.

\textbf{Training Details.} We implement our framework network through pytorch 1.8.1. Follow the settings of Transreid, the training images are augmented with random horizontal, flipping, padding, random cropping and random erasing\cite{30}. Input person images are resized to 256×128, the batchsize is set as 48 , and SGD optimizer is employed with a momentum of 0.9. The weight decay of 1e-4, and the learning rate is initialized as 0.008 with cosine learning rate decay.

\textbf{Evaluation Metrics.} We utilize standard metrics as in
most person ReID literatures, namely Cumulative Match-
ing Characteristic curves (CMC) and mean average precision (mAP), to evaluate the quality of different person re-
identification models. All the experiments are performed in
single query setting.

\subsection{Experimental Results}
\begin{table}[h]
\begin{tabular}{l|cccc}
\hline
               & \multicolumn{4}{c}{Occluded-Duke} \\
Methods        & Rank-1  & Rank-5 & Rank-10 & mAP  \\ \hline
Part Aligned\cite{31}   & 28.8    & 44.6   & 51.0    & 20.2 \\
PCB\cite{12}            & 42.6    & 57.1   & 62.9    & 33.7 \\
Adver Occluded\cite{32} & 44.5    & -      & -       & 32.2 \\ \hline
Part Bilinear\cite{33}  & 36.9    & -      & -       & -    \\
FD-GAN\cite{34}         & 40.8    & -      & -       & -    \\
PGFA\cite{9}           & 51.4    & 68.6   & 74.9    & 37.3 \\
HONet\cite{19}          & 55.1    & -      & -       & 43.8 \\ \hline
DSR\cite{35}            & 40.8    & 58.2   & 65.2    & 30.4 \\
SFR\cite{36}            & 42.3    & 60.3   & 67.3    & 32.0 \\
MoS\cite{37}            & 61.0    & 74.4   & 79.1    & 49.2 \\ \hline
TransReID\cite{21}      & 66.4    & -      & -       & 59.2 \\
DRL-Net\cite{38}        & 65.0    & 79.3   & 83.6    & 50.8 \\
PFT(Ours)      & \textbf{69.8}    & \textbf{83.4}   & \textbf{87.7}    & \textbf{60.8} \\ \hline
\end{tabular}

\caption{Comparison with state-of-the-arts on Occluded-Duke, where PFT shows its better performance than all other methods.}

\end{table}
\textbf{Results on Occluded-Duke Datasets.}  Experimental results on dataset Occluded-Duke are shown in Table 1. We compare with four types of mainstream methods in occluded person re-identification, such as pure holistic re-identification methods (Part Aligned, PCB, Adver Occluded), occluded re-identification method using external semantic information (Part Bilinear, FD-GAN, PGFA, HONet), partial matching method (DSR, SFR, MoS) and transformer method (TransReID, DRL-Net\cite{38}). It can be seen that the transformer based long sequence person re-identification is essentially based on the global feature correlation, and the local features correlation of patch sequences is not well utilized in transformer networks. Therefore to improve the local features correlation in long sequences, our proposed PFT enables transformer to pay attention to local features, and then achieves rank-1 of 69.8\% and mAP of 60.8\%, which performs best on occluded-duke.

\textbf{Results on Occluded-REID and Partial-REID.}
The occluded person images in the Occluded-Reid dataset are much less than the Occluded-Duke dataset. Therefore, many people use the market1501 dataset for pre-training, and then use the Occluded-Reid dataset for testing, so as to achieve a more convergent effect. Here, We choose to use the occluded-Duke dataset for pre-training to achieve better results, since the occluded-Reid dataset is more inclined to the occlusion type. Experimental results verify this hypothesis we put forward, as shown in Table 2.
\begin{table}[h]
\begin{tabular}{l|cc|cc}
\hline
              & \multicolumn{2}{l|}{Occluded-REID}                    & \multicolumn{2}{l}{Partial-REID}                     \\
Methods       & \multicolumn{1}{l}{Rank-1} & \multicolumn{1}{l|}{mAP} & \multicolumn{1}{l}{Rank-1} & \multicolumn{1}{l}{mAP} \\ \hline
PCB\cite{12}           & 41.3                       & 38.9                     & 66.3                       & 63.8                    \\
Part Bilinear\cite{33} & 54.9                       & 50.3                     & 57.7                       & 59.3                    \\
DSR\cite{35}           & 72.8                       & 62.8                     & 43.0                       & -                       \\
FPR\cite{39}           & 78.3                       & 68.0                     & 81.0                       & -                       \\
HOReID\cite{19}        & 80.3                       & 70.2                     & 85.3                       & -                       \\
PVPM\cite{22}          & 70.4                       & 61.2                     & 78.3                       & 72.3                    \\
PGFA\cite{9}          & -                          & -                        & 68.0                       & -                       \\
PFT(OURS)     & \textbf{83.0}              & \textbf{78.3}            & 81.3                       & \textbf{79.9}           \\ \hline
\end{tabular}
\caption{Comparison with state-of-the-arts on Occluded-REID and Partial-REID, where mAP and rank-1 of our methods are  the best results of mainstream methods.}
\label{table2}
\end{table}

The difference between Partial-Reid dataset and Occluded-Reid dataset is that the former focuses on the recognition of some parts of human body, such as arm, upper body and left body, and the former rarely contains occlusion information and other background or noise information. Similar to the Occluded-Duke dataset, images in Occluded-Reid dataset contain more occluded objects and other noise information than Partial-Reid. 

It can be seen from Table 2 that PFT is more suitable to solve the occluded re-identification problem, and its mAP is much higher than the holistic identification method (PCB) and external information method (HOREID). Furthermore, rank-1 of PFT also reached the best, 2.7\% higher than the highest method HOREID.
PFT also shows good performance in Partial-REID dataset, gives full play to the advantages of transformer, and has high overall identification accuracy, so it improves the mAP greatly.

\textbf{Results on Holistic Datasets.} Transreid has demonstrated the powerful performance of transformer as a backbone in the field of person re-identification. Therefore, we hope that our proposed PFT can not only have the best performance for occluded targets, but also show strong generalization ability and robustness in the holistic dataset.

Generally, network backbones are divided into two categories: CNN (PCB, PGFA, VPM, MGCAN, SPReID, OSNet, HOReID, ISP) and transformer (TransReID, DRL-Net). It can be seen from Table 3 that the transformer based method is easier to achieve good results on the holistic Reid dataset. 
\begin{table}[h]
\begin{tabular}{l|cc|cc}
\hline
          & \multicolumn{2}{c|}{Market-1501}                      & \multicolumn{2}{c}{DukeMTMC}                         \\
Methods   & \multicolumn{1}{l}{Rank-1} & \multicolumn{1}{l|}{mAP} & \multicolumn{1}{l}{Rank-1} & \multicolumn{1}{l}{mAP} \\ \hline
PCB\cite{12}       & 92.3                       & 77.4                     & 81.8                       & 66.1                    \\
PGFA\cite{9} & 91.2                       & 76.8                     & 82.6                       & 65.5                    \\
VPM\cite{40}       & 93.0                       & 80.8                     & 83.6                       & 72.6                    \\
MGCAM\cite{41}     & 83.8                       & 74.3                     & 46.7                       & 46.0                    \\
SPReID\cite{42}    & 92.5                       & 81.3                     & -                          & -                       \\
OSNet\cite{43}     & 91.3                       & 84.9                     & 88.6                       & 73.5                    \\
HOReID\cite{19}    & 94.2                       & 84.9                     & 86.9                       & 75.6                    \\
ISP\cite{44}       & 95.3                       & 88.6                     & 89.6                       & 80.0                    \\ \hline
TransReID\cite{21} & 95.2                       & 88.9                     & 90.7                       & 82.0                    \\
DRL-Net\cite{38}   & 94.7                       & 86.9                     & 88.1                       & 76.6                    \\ \hline
PFT(ours) & 95.3                       & 88.8                     & 90.7       
   & 82.1    \\ \hline
\end{tabular}
\caption{Comparison with state-of-the-arts on Market-1501 and DukeMTMC, where the mAP and rank-1 of our method are higher than the traditional mainstream methods.}

\end{table}

Our proposed PFT for occluded person re-identification also works well in the holistic Reid problem. The Rank-1 score and mAP on market-1501 are 1.1\% and 3.9\% higher than the traditional CNN method HOReID respectively. Compared with Transreid method focusing on holistic person re-identification, the performance of these two methods is very close. This demonstrates that PFT can deal with various types of pedestrian re recognition problems.

\textbf{Results on Partial-iLIDS Datasets.} Partial-iLIDS is based on the iLIDS dataset and contains a total of 238 images from 119 people captured by multiple non-overlapping cameras in the airport, and their occluded regions are manually cropped. Because the Partial-iLIDS dataset is too small and contains too few images, we choose other mainstream datasets as the training dataset.

First, we choose to use Occluded-Duke as the training dataset and Partial-iLIDS as the test dataset. It can be seen from Table 4 that our PFT achieves 74.8\% Rank-1 accuracy and 87.3\% rank-3 accuracy. This result is close to the most advanced method at present.
Only with vision transformer, Baseline achieves 71.4\% rank-1 and 87.4\% rank-3 on Partial-iLIDS dataset, which indicates the performance of Baseline is close to HOReID.
Compared with Baseline, our PFT imporves rank-1 by 3.4\%, which demonstrates that the performance of PFT is obviously better than vision transformer.

Second, we want to explore how the performance of PFT will change if we use market1501 as the training dataset.
It can be seen from Table 5 that Baseline still achieves good performance, and rank-1 reaches 73.1\%. Compared with Baseline, the performance of PFT decreases significantly. That is, Rank-1 is only 68.1\%, 6.7\% lower than PFT with Occluded-Duke as the training dataset. We will explore the reasons for this dramatic change in the next section.

\begin{table}[h]
\centering
\begin{tabular}{l|clcllllll}
\cline{1-4}
          & \multicolumn{3}{c}{Partial-iLIDS} \\
Method    & Rank-1       &       & Rank-3        \\ \cline{1-4}
MTRC\cite{45}      & 17.7\%       &       & 26.1\%        \\
AMC+SWM\cite{29}   & 21.0\%       &       & 32.8\%       \\
DSR\cite{35}       & 58.8\%       &       & 67.2\%        \\
SFR\cite{36}       & 63.9\%       &       & 74.8\%        \\
VPM\cite{40}       & 65.5\%       &       & 74.8\%        \\
FPR\cite{39}       & 68.1\%       &       & -             \\
PGFA\cite{9}      & 69.1\%       &       & 80.9\%        \\
HOReID\cite{19}    & 72.6\%       &       & 86.4\%        \\
MHSA-Net\cite{46}  & 74.9\%       &       & 87.2\%      \\ \cline{1-4}
Baseline  & 71.4\%       &       & 87.4\%        \\
PFT(Ours) & 74.8\%       &       & 87.3\%       \\ \cline{1-4}
\end{tabular}
\caption{Comparison with state-of-the-arts method on Partial-iLIDS, where Occluded-Duke is utilized as the training dataset and Baseline refers to the original vision transformer network.}
\label{table4}
\end{table}

\subsection{Analysis of different training dataset}
It can be seen from Table 4 and Table 5 that different training dataset will change the performance of our PFT greatly on the same test dataset. With market1501 as the training set and vision transformer as the network backbone, rank-1 and rank-3 achieve a high level.
\begin{table}[h]
\centering
\begin{tabular}{l|clcllllll}
\cline{1-4}
          & \multicolumn{3}{c}{Partial-iLIDS} \\
Method    & Rank-1       &       & Rank-3        \\ \cline{1-4}
MTRC\cite{45}      & 17.7\%       &       & 26.1\%        \\
AMC+SWM\cite{29}   & 21.0\%       &       & 32.8\%       \\
DSR\cite{35}       & 58.8\%       &       & 67.2\%        \\
SFR\cite{36}       & 63.9\%       &       & 74.8\%        \\
VPM\cite{40}       & 65.5\%       &       & 74.8\%        \\
FPR\cite{39}       & 68.1\%       &       & -             \\
PGFA\cite{9}      & 69.1\%       &       & 80.9\%        \\
HOReID\cite{19}    & 72.6\%       &       & 86.4\%        \\
MHSA-Net\cite{46}  & 74.9\%       &       & 87.2\%      \\ \cline{1-4}
baseline  & 73.1\%       &       & 84.0\%        \\
PFT(Ours) & 68.1\%       &       & 81.5\%       \\ \cline{1-4}
\end{tabular}
\caption{Comparison with state-of-the-arts method on Partial-iLIDS, where market1501 is utilized as the training dataset and Baseline refers to the original vision transformer network.}
\label{table5}
\end{table}

\begin{figure}[h]
    \centering
    \includegraphics[width=1.0\linewidth]{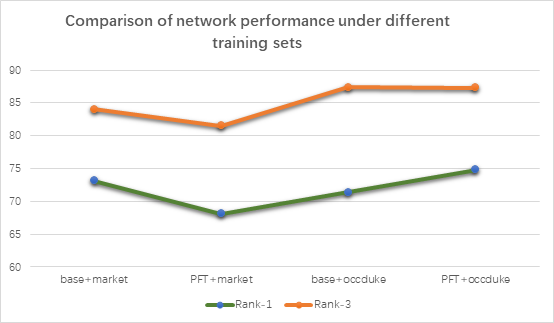}
    \caption{Performance analysis of different training dataset} 
    \label{fig:comparison}
\end{figure}

As shown in Figure \ref{fig:comparison}, rank-1 of Baseline reaches 73.1\%, only 1\% lower than the most advanced method.
For the overall long-range sequence correlation, vision transformer has powerful generalization ability and achieves excellent results for small dataset like Partial-iLIDS.
Taking the proposed PFT as the network backbone, it is difficult to achieve satisfactory results on Partial-iLIDS when using market1501 as the training dataset. On the contrary, with occluded-Duke dataset as the training set, our PFT has reached the best level at present. That is, rank-1 reaches 74.8\%, almost equal to the performance of MHSA-Net. Compared with baseline using Occluded-Duke as the training dataset, the performance of PFT is improved by 3.4\%.

The above description shows that PFT is indeed effective and better than traditional vision transformer network in the field of occluded person re-identification. 
For baseline, its rank-1 with Occluded-Duke as traning dataset is 1.7\% lower than that with market1501 as traning dataset, which demonstrates that the performance of Vision transformer degrades when dealing with occluded person re-identification. On the contrary, our PFT achieves better results under the same conditions. The above analysis clarifies that PFT is better at solving complex feature extraction problems, such as occluded person re-identification.

\subsection{Ablation study}
In this section, we will research the effectiveness of each module of the proposed PFT. We use vision transformer as the baseline and conduct ablation experiments of PFDE (Patch Full Dimension Enhancement), FRM (Fusion and Reconstruction Module) and SSM (Spatial Slicing Module). Occluded-Duke is here selected as the target dataset to verify the effectiveness of proposed different modules in re-identification of occluded people. The ablation study results are shown in Table 6.
\begin{table}[h]
\begin{tabular}{l|llll|llll}
\hline

Index & $B$ & $P$ & $F$ & $S$ & R-1  & R-5 & R-10 & mAP  \\ \hline
1     & $\surd$ &   &   &   & 60.6 &77.1     &81.9      & 53.1 \\
2     & $\surd$ & $\surd$ &   &   & 62.8 &78.2     &83.8      & 55.0 \\
3     & $\surd$ & $\surd$ & $\surd$ &   & 64.6 &80.1     &85.2      & 56.9 \\
4     & $\surd$ &  & $\surd$ & $\surd$ & 67.2 &81.4     &86.2      & 58.3   \\
5     & $\surd$ &$\surd$ &  & $\surd$ & 67.3 &82.0     &87.1      & 59.0    \\ 
6     & $\surd$ & $\surd$ & $\surd$ & $\surd$ & 69.8 &83.4     &87.7      & 60.8 \\ \hline
\end{tabular}
\caption{Results of ablation study for PFT, where $B$ represents vision transformer baseline, $P$ represents PFDE, $F$ represents FRM and $S$ represents SSM.}
\label{table6}

\end{table}

{\bf Effectiveness of Patch Full Dimension Enhancement Module.}
First, we only utilize PFDE based on the vision transformer baseline. In this way, the PFDE module can optimize the construction of patch sequence and extract discriminative features. Furthermore, the full dimension enhancement of patch sequence can enrich the diversity of data samples, improve the differentiation of data samples, enlarge the fine categorization distance within a reasonable range and prevent falling into local optimization by deeply embedding the patch sequence tensor. As shown in index 2 of Table 6, compared with baseline, the rank-1 is improved by 2.2\%. This shows that the full dimension enhancement module does play a role in the occlusion problem with complex feature information.

In order to verify the wide applicability of PFDE, we compared baseline + FRM + SSM with baseline + FRM + SSM + PFDE. These two networks add two other modules on the basis of vision transformer, and the flow of data samples in the network is more complex.
It can be seen from index numbers 4 and 6, rank-1 of index 4 is significantly improved by 2.5\% than index 6.
In other words, the enhancement effect of PFDE module on patch sequence is proved to be feasible, and PFDE module has good generalization performance in different situations such as complex vision transformer and original vision transformer. It can bring a positive effect on occluded person re-identification, and will not have a negative impact on network.

{\bf Effectiveness of Fusion and Reconstruction Module.}
The FRM module focuses on the reconstruction of patch sequence, therefore it can be easily inserted into the transformer network.
It can be seen from index 2 and index 3 that after adding FRM module, rank-1 and mAP are improved by 3\% and 2\% respectively.
Through the comparison of index 5 and 6, after adding the space slicing module, FRM module still shows its effectiveness in solving occluded problem, by providing an additional + 2\% rank-1 score and + 1.6\% mAP.

Through the above two groups of comparative experimental results, it can be found that FRM is feasible to reduce the noise in occluded re-identification. While weakening the occluded feature information, FRM can extract more discriminative feature information, which increases the proportion of occluded person distinguishable feature information in the newly constructed feature patch sequence. 
In other words, the patch sequence output from the FRM module will contain more feature information related to occluded person, which will undoubtedly help to further improve the network performance of the vision transformer network.

{\bf Effectiveness of Spatial Slicing Module.} 
SSM module can integrate the spatial correlation of the image into the patch sequence to improve the short-range correlation of the patch sequence, thereby enhancing the network's ability to extract local features of occluded person.

In index 3, baseline+PFDE+FRM achieves rank-1 of 64.6\% and mAP of 56.9\%. Compared with index 3, SSM module is added to index 6. It can be seen from index 3 and index 6 that the experimental effect has been greatly improved, where rank-1 increased by 5.2\% and mAP increased by 3.9\%.

It can be seen from index 4, 5 and 6 that the network performance on occluded re-identification will be greatly improved, which indicates that the spatial correlation in patch sequence of vision transformer can be used reasonably with the addition of SSM module.

Therefore, spatial correlation of input images helps to improve visual selective attention to local features. For occluded person re-identification, the integrated spatial correlation in patch sequence can expand the difference between similar persons, which is effective for complex fine categorization problems.

\subsection{Analysis of Initial Strengthening Coefficient}
Several groups of comparative experiments are conducted to evaluate the effect of initial strengthening coefficient on the performance of PFDE module. 

The learnable patch dimension enhancement tensor $LPDE$ is initialized into Gaussian distribution, uniform distribution, Laplace distribution and exponential distribution respectively.
Experimental results demonstrate that different distributions will change the original feature distribution of the image to varying degrees. It is difficult to correct the feature deviation back to the original distribution through only one layer of PFDE, which will result in the local optimization of vision transformer network.
Therefore, the learnable patch embedding $LPDE$ is initialized to
a tensor filled with the scalar value 1. Namely, the image feature distribution is not adjusted.

\begin{figure}[h]
    \centering
    \includegraphics[width=1.0\linewidth]{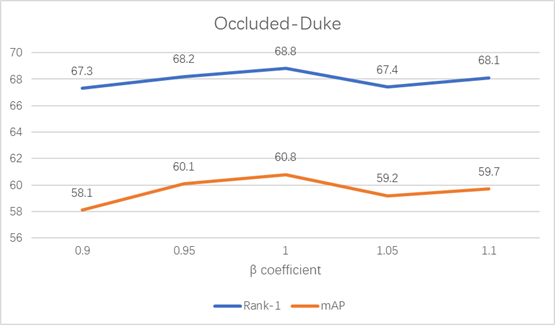}
    \caption{Analysis of  Initial strengthening coefficient $\beta$} 
    \label{fig:Overview1}
\end{figure}

According to experimental results as shown in Figure \ref{fig:Overview1}, different initial strengthening coefficient $\beta $ can affect the convergence speed and convergence accuracy of vision transformer network. 
Specifically, it can be clearly seen that the change of $\beta$ will have an impact on the network performance. Only with $\beta$ being 1.0, all aspects of network performance can reach the best. This demonstrates that PFDE has the ability to optimize the network within a certain range. Once the value of $\beta$is outside the range [0.95-1.05], the performance of the network will fluctuate greatly, and it is difficult for PFDE to play its optimization ability.

According to the experimental results, the following conclusions can also be obtained. When the original feature distribution of the image changes greatly, PFDE can still maintain good performance. Namely, both rank-1 and mAP achieve a high level, which also denotes the effectiveness of PFDE and PFT.

\section{Conclusion}
This paper proposes a PFT network based on vision transformer, which includes three newly designed modules. 
SSM in PFT can make good use of the spatial information of input image and integrate it into patch sequence, so as to improve the short-range correlation of patch sequence and focus on locally distinguishable features of occluded person. PFDE module can enrich the diversity of input images and optimize the distribution of feature map, so as to improve the generalization and robustness of the network. FRM module can highlight the distinguishable characteristics of occluded person and desalinate the noise information.
Extensive experiments on occluded, partial and holistic datasets demonstrate the effectiveness of
our proposed PFT framework.

\section{Acknowledgement}
This work is supported by Natural Science Foundation of Nanjing University of Posts and Telecommunications under No. NY219107, and National Natural Science Foundation of China under No. 52170001.





\bibliographystyle{ieeetr}

\end{document}